\pdfoutput=1

\documentclass[11pt]{article}

\usepackage{emnlp2021}

\usepackage{times}
\usepackage{latexsym}

\usepackage[T1]{fontenc}

\usepackage[utf8]{inputenc}

\usepackage{microtype}

\usepackage{graphicx}
\usepackage{todonotes}
\usepackage{subcaption}
\usepackage{makecell}
\usepackage{multirow}
\usepackage{todonotes}
\usepackage{amsmath}
\usepackage{enumitem}
\usepackage{hyperref}
\usepackage{pgfplots}
\usepackage{caption}
\usepackage{algorithm}
\usepackage{algpseudocode}
\usepackage{tabularx,ragged2e}
\usepackage{pifont}
\usepackage{setspace}
\usepackage{float}

\title{Effective Sequence-to-Sequence Dialogue State Tracking}

\author{Jeffrey Zhao, Mahdis Mahdieh, Ye Zhang, Yuan Cao, Yonghui Wu 
\\
Google Research\\
\texttt{\{jeffreyzhao,mahdis,yezhan,yuancao,yonghui\}@google.com} \\
}

\begin{document}
\maketitle
\begin{abstract}
Sequence-to-sequence models have been applied to a wide variety of NLP tasks, but how to properly use them for dialogue state tracking has not been systematically investigated. In this paper, we study this problem from the perspectives of pre-training objectives as well as the formats of context representations. We demonstrate that the choice of pre-training objective makes a significant difference to the state tracking quality. In particular, we find that masked span prediction is more effective than auto-regressive language modeling. We also explore using Pegasus, a span prediction-based pre-training objective for text summarization, for the state tracking model. We found that pre-training for the seemingly distant summarization task works surprisingly well for dialogue state tracking. In addition, we found that while recurrent state context representation works also reasonably well, the model may have a hard time recovering from earlier mistakes. We conducted experiments on the MultiWOZ 2.1-2.4, WOZ 2.0, and DSTC2 datasets with consistent observations.
\end{abstract}

\section{Introduction}

Sequence-to-sequence (Seq2Seq) modeling \cite{ilya14sequence} is one of the most widely adopted generative framework for a multitude of NLP tasks. While it has also been applied for task-oriented dialogue modeling \cite{wen-etal-2018-sequence,lei-etal-2018-sequicity,subendhu20parse,feng2020sequencetosequence}, how to best setup Seq2Seq models on this task remains an understudied topic. In this paper, we investigate this problem from two perspectives: Pre-training objectives and dialogue context representation, and we focus on the dialogue state tracking (DST) task.

The flexibility of the Seq2Seq model allows us to adopt and compare pre-training strategies for other NLP tasks sharing the same architecture. Specifically, we first experimented with different pre-training setups of T5 \cite{raffel20exploring} which have been shown to be effective for generic language understanding. Additionally, as an exploratory effort, we applied Pegasus \cite{zhang20ae}, a pre-training procedure designed for text summarization, to the task of DST.

Additionally, to investigate how different dialogue context representations affect Seq2Seq performance, we compare two versions of all models: one accepts full conversation history as context, and one that feeds the previously predicted states recurrently as summary of context.

We conduct systematic experiments on the MultiWOZ \cite{budzianowski-etal-2018-multiwoz} benchmark. For fair comparison with existing approaches, we report results on MultiWOZ 2.1-2.4 \cite{eric2019multiwoz,zhang2020multiwoz,han2020multiwoz,ye2021multiwoz}, all 4 variations of the benchmark proposed to date. In addition, we report results on the WOZ 2.0 \cite{wen2016networkbased} and DSTC2 \cite{henderson2014second} datasets. Our findings can be summarized as follows:
\begin{enumerate} [nolistsep]
    \item Pre-training procedures involving masked span prediction work consistently better than auto-regressive language modeling objectives.
    \item Pre-training for text summarization works surprisingly well for DST, despite it being a seemingly irrelevant task.
    \item Recurrent models work reasonably well by including previously predicted states and constant length dialogue history. However they may suffer from the problem of not being able to recover from early mistakes.
\end{enumerate}

\section{Methods}
\subsection{Models} \label{sec:models}
We directly apply the Seq2Seq model to the problem of state tracking, where both the encoder and decoder are Transformers \cite{vaswani17attention}. The inputs to the encoder are dialogue contexts, and the decoder generates a sequence of strings of the format \texttt{slot1=value1,slot2=value2,...} describing the predicted states conditioned on the given context. Depending on how we represent the dialogue context, we consider two variations of the model:

\begin{enumerate}
    \item \textbf{Full-history model}: The most straightforward way of preparing the context is simply to concatenate turns from the entire history as inputs to the encoder, which ensures the model to have full access to the raw information required to predict the current state. This setup is also adopted by several generative dialogue models such as SimpleTOD \cite{hosseiniasl2020simple}, Seq2Seq-DU \cite{feng2020sequencetosequence} and SOLOIST \cite{peng2021soloist}. A potential drawback of the full-history model is that it may become increasingly inefficient as a conversation unfolds and the input length grows.

\item \textbf{Recurrent-state model}: An alternative approach is to include just the $N$ recent turns in the conversation history, and replace turns from 1 to $T-N$ with dialogue states up to $T-N$ (where $T$ is the current turn index). That is, the inputs to encoder have the format $[\texttt{states}(turn_{1,\ldots,T-N}), turn_{T-N+1,\ldots,T}]$. States provide a summarization of the conversation semantics. By consolidating remote histories into states we not only reduce the context lengths, but also discard information not immediately related to the purpose of state tracking. Similar setup has also been considered by generative models including GPT-2 \cite{budzianowski-vulic-2019-hello} and Sequicity \cite{lei-etal-2018-sequicity}, although in their cases only the last turn has been considered ($N=1$).
\end{enumerate}

An example of the input and output formats for both models is given in Appendix \ref{sec:appendix-dialog-example}.

\subsection{Pre-training}
Pre-training followed by task-specific fine-tuning is becoming a standard paradigm for contemporary NLP model training. Existing pre-training objectives mainly fall into two categories: masked span prediction (where the span length can be 1 corresponding to word prediction) and auto-regressive prediction. Objectives like BERT MLM \cite{devlin-etal-2019-bert} and the denoising setup in T5 \cite{raffel20exploring} belong to the former category, while GPT \cite{radford2019language,NEURIPS2020_1457c0d6} and the prefix LM setup in T5 fall into the latter.

For generative dialogue modeling, both pre-training styles have been considered. For example, Seq2Seq-DU \cite{feng2020sequencetosequence} adopted a BERT-pre-trained encoder, while SimpleTOD \cite{hosseiniasl2020simple} and SOLOIST \cite{peng2021soloist} are based on the GPT-2 auto-regressive prediction procedure. Nevertheless, it remains unclear which style is more effective for dialogue understanding. To study this problem, we compare span prediction and auto-regressive language modeling (ARLM) by pre-training the encoder and decoder simultaneously using the denoising and prefix LM objectives from T5. To compare the relative effectiveness of different pre-training styles, we consider 3 setups: 1) Pre-training the model with span prediction only; 2) Continuing the pre-training  of models from setup (1) with prefix LM; 3) Pre-training the model only with prefix LM only.

While T5 pre-training has demonstrated its effectiveness for generic language understanding tasks such as the GLUE and SuperGLUE benchmarks, we are curious as to which procedures are biased towards the downstream DST task. While it can be difficult to define an objective that applies immediately to DST, we consider a surrogate pre-training for a seemingly remote task: Summarization. To properly summarize a large chunk of text requires the model to be able to extract key semantics out of a clutter of inputs, which to some extent shares a similar problem structure as DST.

Following this intuition, we choose Pegasus \cite{zhang20ae}, a strong pre-training objective developed for summarization based on Seq2Seq, as an alternative for comparison. In brief, Pegasus defines a self-supervised objective named Gap Sentence Generation (GSG), which identifies potentially important sentences in a paragraph according to some heuristics (for example, the top-$m$ sentences with the highest ROUGE score with respect to the remaining ones), masks them out, and forces the decoder to predict these pivoting sentences. A critical difference between Pegasus and other span prediction objectives is that masked spans are carefully identified instead of randomized. This principled operation positions the model to work specifically well for the downstream task of summarization.

\section{Experiments}

\subsection{General Setup}

Our models are built with the open-source framework Lingvo \cite{shen2019lingvo}\footnote{\url{https://github.com/tensorflow/lingvo}}. Each encoder and decoder has 12 Transformer layers, 8 attention head's and embedding dimension 768. Our models are trained with 16 TPUv3 chips \cite{jouppi17indata}. We use the memory-efficient Adafactor \cite{pmlr-v80-shazeer18a} as the optimizer, with learning rate 0.01 and inverse squared root decay schedule. We use the default SentencePiece model provided by T5\footnote{\url{https://github.com/google-research/text-to-text-transfer-transformer}} with vocabulary size 32k. For the pre-training procedure, we strictly follow the setups and procedures described in \cite{zhang20ae} and \cite{raffel20exploring}.
For decoding, we use beam search with size 5. We also enabled label smoothing with uncertainty 0.1 during training.

\subsection{Datasets}

We conduct our experiments on the MultiWOZ \cite{budzianowski-etal-2018-multiwoz} benchmark. The original MultiWOZ dataset, released in 2018, was known to contain substantial annotation errors. Continuous efforts have been made in recent years to clean up and refine the annotations, resulting in 4 variations of the dataset (2.1-2.4, \citet{eric2019multiwoz,zhang2020multiwoz,han2020multiwoz,ye2021multiwoz}). The existence of multiple versions of the same benchmark, as well as ad-hoc pre- and post-processing procedures\footnote{For example on MultiWOZ 2.1, some well-known works including TRADE \cite{wu-etal-2019-transferable}, TripPy \cite{heck-etal-2020-trippy} and SimpleTOD \cite{hosseiniasl2020simple} applied different data processing procedures, making the results incomparable.} adopted by different research groups make it difficult to compare results fairly. We therefore report results on all of MultiWOZ 2.1-2.4\footnote{MultiWOZ datasets retrieved from: \url{https://github.com/budzianowski/multiwoz} (2.1, 2.2), \url{https://github.com/lexmen318/MultiWOZ-coref} (2.3), \url{https://github.com/smartyfh/MultiWOZ2.4} (2.4)},
\emph{without} any pre- or post-processing of the original data. We use Joint-Goal-Accuracy (JGA) as the metric for all experiments.

In addition to the MultiWOZ datasets, we also report results on the WOZ 2.0\footnote{WOZ 2.0 dataset retrieved from \url{https://github.com/nmrksic/neural-belief-tracker/tree/master/data/woz}} \cite{wen2016networkbased} and DSTC2\footnote{DSTC2 dataset retrieved from \url{https://github.com/matthen/dstc}} \cite{henderson2014second} datasets. While these datasets are much smaller in both ontology and number of examples when compared to MultiWOZ, they provide additional evidence for the conclusions we make in this paper.

We compare our results with a set of strong baselines: TRADE \cite{wu-etal-2019-transferable}, SUMBT \cite{lee-etal-2019-sumbt}, DS-DST \cite{zhang2020classify}, Seq2Seq-DU \cite{feng2020sequencetosequence}, SOM-DST \cite{kim-etal-2020-efficient}, Transformer-DST \cite{zeng2021jointly}, TripPy \cite{heck-etal-2020-trippy}, SAVN \cite{wang-etal-2020-slot}, SimpleTOD \cite{hosseiniasl2020simple}, StateNet \cite{ren2018statenet}, GLAD \cite{zhong2018glad}, GCE \cite{nouri2018gce}, and Neural Belief Tracker \cite{mrksic2016nbt}. To be consistent with our approach, when both open- and closed-vocabulary setups are available, we only compare with the open-vocabulary setup.

Note that on DSTC2, unlike other methods which combines the $n$-best speech recognition hypotheses as inputs, we make use of only the top 1-best hypothesis for simplicity, although the combination of $n$-best hypotheses could potentially further improve DST quality.

\subsection{Results}\label{sec:results}
We first report the MultiWOZ JGA scores achieved by the full-history models in Table \ref{table:Seq2Seq_full}, in which ``span'' and ``ARLM'' indicate masked span prediction and auto-regressive language modeling for pre-training respectively, and ``span+ARLM'' means pre-training with ``span'' followed by ``ARLM''. In addition, WOZ and DSTC2 JGA scores are reported in Table \ref{table:seq2seq_woz_dstc}.
\begin{table}[htbp]
\small
\centering
    \centering
    \scalebox{0.95}{
    \begin{tabular}{lcccc}
    \hline
    & \multicolumn{4}{c}{\textbf{MultiWOZ}} \\
    \textbf{Model} & \textbf{2.1}\ding{72} & \textbf{2.2} & \textbf{2.3} & \textbf{2.4} \\
    \hline
    TRADE & 45.6 & 45.4 & 49.2 & 55.1\\
    SUMBT & 49.2 & 49.7 & 52.9 & 61.9 \\
    DS-DST & 51.2 & 51.7 & - & - \\
    Seq2Seq-DU & - & \textbf{54.4} & - & - \\
    Transformer-DST & \textbf{55.35} & - & - & - \\
    SOM-DST & 51.2 & - & 55.5 & \textbf{66.8}\\
    TripPy & 55.3 & - & \textbf{63.0} & 59.6\\
    SAVN & 54.5 & - & 58.0 & 60.1\\
    SimpleTOD\ding{115} & 50.3/\textbf{55.7} & - & 51.3 & -\\
    \hline
    Pegasus & \textbf{54.4} & 56.6 & \textbf{60.2} & 66.6 \\
    T5 (span) & 52.8 & \textbf{57.6} & 59.3 & \textbf{67.1}\\
    T5 (span+ARLM) & 53.1 & 57.1 & 59.9 & 65.6 \\
    T5 (ARLM) & 52.5 & 56.1 & 58.9 & 63.0\\
    No pre-train & 25.8 & 26.1 & 28.1 & 26.7 \\
    \hline
    \end{tabular}
    }
    \caption{JGA comparison on MultiWOZ 2.1-2.4 with the full history model. \ding{72}: For 2.1, baseline methods adopted different and incomparable data clean-up procedures, but we used the original data and did not do any pre- or post-processing for convenient and fair future comparisons. \ding{115}: SimpleTOD results are cited from the 2.3 website \url{https://github.com/lexmen318/MultiWOZ-coref}, in which two numbers are reported for 2.1 (one produced by the 2.3 author, the other by the original SimpleTOD paper). ``-'' indicates no public number is available. Best results given by existing and our models are marked in bold.}
    \label{table:Seq2Seq_full}
\end{table}

\begin{table}[htbp]
\small
\centering
    \centering
    \scalebox{0.95}{
    \begin{tabular}{lcccc}
    \hline
    \textbf{Model} & \textbf{WOZ 2.0} & \textbf{DSTC2} \\
    \hline
    SUMBT & \textbf{91.0} & - \\
    StateNet-PSI & 88.9 & \textbf{75.5} \\
    GLAD & 88.1 & 74.5 \\
    GCE & 88.5 & - \\
    Neural Belief Tracker: NBT-DNN & 84.4 & 72.6 \\
    Neural Belief Tracker: NBT-CNN & 84.2 & 73.4 \\
    \hline
    Pegasus & \textbf{91.0} & \textbf{73.6} \\
    T5 (span) & \textbf{91.0} & \textbf{73.6} \\
    T5 (span+ARLM) & \textbf{91.0} & 73.5 \\
    T5 (ARLM) & 89.5 & 73.3 \\
    No pre-train & 64.5 & 50.1 \\
    \hline
    \end{tabular}
    }
    \caption{JGA comparison for WOZ 2.0 and DSTC2 datasets on the full history model. Note that our DSTC2 JGAs are likely underreported. While other models use the n-best predictions to evaluate, we only used the single best prediction.}
    \label{table:seq2seq_woz_dstc}
\end{table}

From Tables \ref{table:Seq2Seq_full} and \ref{table:seq2seq_woz_dstc} we make the following observations:
\begin{enumerate}
  \setlength\itemsep{0.0em}
    \item Pre-training procedures with masked span prediction involved (``span'', ``span+ARLM'') consistently performed better than using ``ARLM'' alone. This is true even if ``span'' is continued by ``ARLM'', and this result is seen in not just MultiWOZ 2.1-2.4 but also WOZ 2.0 and DSTC2.
    \item Pegasus pre-training works almost equally well or better than the T5 pretraining, indicating that some features can be shared and transferred between the two tasks. Again, this observation is consistent across all benchmarks. This also corroborates conclusion (1) above in that span prediction objectives are more effective for DST.
    \item Without pre-training, model quality drops miserably, as expected.
\end{enumerate}

\subsection{Recurrent Results}\label{sec:recresults}

For the recurrent-state model, we report results for the Pegasus pre-trained model on MultiWOZ 2.1-2.4 in Table \ref{table:Seq2Seq_rec}, with $N=1,2,3$ (number of recent history turns). Each turn contains a pair of user and agent utterance. Our observations on models pre-trained with T5 are similar. The results show that while the recurrent-state models achieved reasonably good JGA on all data sets, they are nevertheless worse than the full-history model, despite the fact that the representation of context is more concise for the recurrent model. What is more, the choice of $N$ can make a big difference to the model quality.

\begin{table}[htbp]
\small
\centering
    \centering
    \scalebox{0.95}{
    \begin{tabular}{lcccc}
    \hline
    \textbf{Model} & \textbf{2.1}\ding{72} & \textbf{2.2} & \textbf{2.3} & \textbf{2.4}\\
    \hline
    Pegasus (full-history) & 54.4 & 56.6 & 60.2 & 66.6 \\
    \hline
    Pegasus (1-turn history) &  \makecell{52.4} &  \makecell{53.9} &  \makecell{\textbf{59.5}} &  \makecell{57.0}\\
    Pegasus (2-turn history) &  \makecell{\textbf{52.7}} &  \makecell{\textbf{56.2}} &  \makecell{59.2} &  \makecell{59.3}\\
    Pegasus (3-turn history) &  \makecell{52.3} &  \makecell{55.9} &  \makecell{58.4} &  \makecell{\textbf{60.0}}\\
    \hline
    \end{tabular}
    }
    \caption{JGA of the recurrent model pre-trained with Pegasus.
    \ding{72} has the same meaning as in Table \ref{table:Seq2Seq_full}.}
    \label{table:Seq2Seq_rec}
\end{table}

A closer look at failed examples produced by the model reveals that the main reason why the recurrent context representation achieved worse results is that they had a hard time recovering from prediction mistakes made at earlier turns. Since previously predicted states are feedback to the model inputs for future predictions, as long as the model made a mistake at earlier turns, this wrong prediction will be carried over as future inputs, causing the model to make consecutive prediction errors. We therefore believe that for the DST task, it may still be important to provide the model full access to dialogue history, so that it can learn to correct its predictions once a mistake was made in the past.

\subsection{Remarks}
From results enumerated in Sec. \ref{sec:results}, one will see wildly varying scores across MultiWOZ 2.1-2.4, despite the fact that each dataset evolved from the same benchmark. This poses a concerning question of whether existing approaches can generalize well across different setups and benchmarks. For example, TripPy performed remarkably well on 2.1 and 2.3 (55.3\%, 63.0\%), but dropped to 59.6\% on 2.4 (which claims to be the ``cleanest'' version of MultiWOZ). SOM-DST on the other hand, under-performed on 2.1 and 2.3 but achieved a strong result on 2.4.

We therefore suggest researchers working on the MultiWOZ benchmark report results on multiple version of the data with consistent or no data processing steps, to provide the community a more faithful assessment of the quality of their approaches.

\section{Related Work}
Generative sequential models have been applied for task-oriented dialogue problems in several ways. \cite{budzianowski-vulic-2019-hello,hosseiniasl2020simple,peng2021soloist} adopted GPT-2, a uni-directional pretrained Transformer LM, as backbones for the generation of states, actions and responses. Under the framework of Seq2Seq, perhaps most similar to our work is \cite{feng2020sequencetosequence}, which adopts a Transformer encoder-decoder architecture, with the encoder pre-trained with BERT which is also used to encode schema. Besides, \cite{wen-etal-2018-sequence} is an early example that uses encoder outputs as state representation, merged with KB representation for the decoder to generate responses; \cite{lei-etal-2018-sequicity} proposes a simplistic two stage CopyNet on top of Seq2Seq model to enable word copying from input sequences; \cite{chen2020credit} proposes a hierarchical Seq2Seq model for coarse-to-fine DST; \cite{zeng2021jointly} proposes a ``flat'' encoder-decoder structure which reuses a BERT encoder for the function of a decoder with hidden layer states reused.

In terms of pre-training, BERT and GPT are still the most commonly used techniques \cite{zaib2021short,zhang-etal-2020-dialogpt}. Various pre-training methods developed for dialogue-specific problems have also been developed. \cite{zhang2021improving} uses dialogue specific datasets for pre-training and fine-tuning. \cite{mehri-etal-2019-pretraining} studies 4 ways of pre-training aiming at better capturing discourse-level dependencies for multi-turn dialogues; \cite{li2020taskspecific} proposed a contrastive pre-training loss to capture important qualities of dialogues; \cite{bao2021unified} proposed a curriculum pre-training procedure for response generation, subsuming open-domain, knowledge-grounded, task-oriented dialogue applications. \cite{liu2021pretraining} factorize the generative dialogue model according to the noisy channel model, pre-training each component separately.

\section{Conclusion}
We studied the problem of how to perform the DST task with Seq2Seq models effectively from the perspective of pre-training and context representation. We demonstrated that Seq2Seq pre-training objectives involving masked span prediction are more preferred than auto-regressive predictions for dialogue understanding. This observation further generalizes to the adoption of Pegasus, a span prediction objective for summarization, which works surprisingly well on DST tasks. We also find that recurrent state representation for dialogue context can work reasonably well.

\bibliography{anthology,emnlp2021}
\bibliographystyle{acl_natbib}

\appendix

\section{Appendix}
\label{sec:appendix}

\subsection{Dialog Example}
\label{sec:appendix-dialog-example}
Dialog examples are formatted to the following input and output sequences for the models presented in this paper. An example input sequence for a full-history model:

\begin{figure}[H]
\setstretch{0.75}
\begin{small}
\texttt{user: I need to find a spot on a train on wednesday, can you help me find one? agent: yes I can. where are you going and what time would like to arrive or depart? user: i'm leaving from london kings cross and going to cambridge. i'd like to leave after 14:30 on wednesday. agent: where would you be departing from? user: i am looking to depart from broxbourne.}
\end{small}
\end{figure}

The target output sequence would be

\begin{figure}[H]
\setstretch{0.75}
\begin{small}
\texttt{train-day = wednesday; train-departure = london kings cross; train-destination = cambridge; train-leaveat = 14:30}
\end{small}
\end{figure}
As a recurrent example, we would remove the older turns of the conversation and replace them with the relevant states. For example, the 2-turn recurrent input sequence for this example would be

\begin{figure}[H]
\setstretch{0.75}
\begin{small}
\texttt{<state> train-day = wednesday <utterance> agent: yes I can. where are you going and what time would like to arrive or depart? user: i'm leaving from london kings cross and going to cambridge. i'd like to leave after 14:30 on wednesday. agent: where would you be departing from? user: i am looking to depart from broxbourne.}
\end{small}
\end{figure}

The 1-turn recurrent input sequence would be
\begin{figure}[H]
\setstretch{0.75}
\begin{small}
\texttt{<state> train-day = wednesday; train-departure = london kings cross; train-destination = cambridge; train-leaveat = 14:30 <utterance> agent: where would you be departing from? user: i am looking to depart from broxbourne.}
\end{small}
\end{figure}

Note that the target output sequence remains the same for the recurrent input sequences. That is, the model is expected to carry over the predictions from previous states into the current state. Empirically, we found this approach works better than only predicting the new states at each turn.

\end{document}